\documentclass[conference]{IEEEtran}

\usepackage{cite}
\usepackage{cuted}
\usepackage{float}
\usepackage{url}
\usepackage{placeins}
\usepackage{graphicx}
\usepackage{tabularx}
\usepackage{multirow}
\usepackage{booktabs}
\usepackage{physics}
\usepackage{amsfonts}

\begin{document}

\title{A Deep Learning Framework for Three Dimensional Shape Reconstruction from Phaseless Acoustic Scattering Far-field Data \\
}

\makeatletter
\newcommand{\linebreakand}{%
  \end{@IEEEauthorhalign}
  \hfill\mbox{}\par
  \mbox{}\hfill\begin{@IEEEauthorhalign}
}
\makeatother

\author{\IEEEauthorblockN{Doga Dikbayir}
\IEEEauthorblockA{\textit{Computer Science and Engineering} \\
\textit{Michigan State University}\\
East Lansing, Michigan, USA \\
dikbayir@msu.edu}
\and
\IEEEauthorblockN{Abdel Alsnayyan}
\IEEEauthorblockA{\textit{Electrical and Computer Engineering} \\
\textit{Michigan State University}\\
East Lansing, Michigan, USA}
\and
\IEEEauthorblockN{Vishnu Naresh Boddeti}
\IEEEauthorblockA{\textit{Computer Science and Engineering} \\
\textit{Michigan State University}\\
East Lansing, Michigan, USA \\
vishnu@msu.edu}
\linebreakand
\IEEEauthorblockN{Balasubramaniam Shanker}
\IEEEauthorblockA{\textit{Electrical and Computer Engineering} \\
\textit{The Ohio State University}\\
Columbus, Ohio, USA \\
shanker@ece.osu.edu}
\and
\IEEEauthorblockN{Hasan Metin Aktulga}
\IEEEauthorblockA{\textit{Computer Science and Engineering} \\
\textit{Michigan State University}\\
East Lansing, Michigan, USA \\
hma@msu.edu}

}

\maketitle

\begin{abstract}
The inverse scattering problem is of critical importance in a number of fields, including medical imaging, sonar, sensing, non-destructive evaluation, and several others. The problem of interest can vary from detecting the shape to the constitutive properties of the obstacle. The challenge in both is that this problem is ill-posed, more so when there is limited information. That said, significant effort has been expended over the years in developing solutions to this problem. Here, we use a different approach, one that is founded on data. Specifically, we develop a deep learning framework for shape reconstruction using limited information with single incident wave, single frequency, and phase-less far-field data. This is done by (a) using a compact probabilistic shape latent space, learned by a 3D variational auto-encoder, and (b) a convolutional neural network trained to map the acoustic scattering information to this shape representation. The proposed framework is evaluated on a synthetic 3D particle dataset, as well as ShapeNet, a popular 3D shape recognition dataset. As demonstrated via a number of results, the proposed method is able to produce accurate reconstructions for large batches of complex scatterer shapes (such as airplanes and automobiles), despite the significant variation present within the data.
\end{abstract}


\section{\label{sec:1} Introduction \& Related Work}

Inverse acoustic scattering problems (IASP) \cite{invscatthe} have been extensively studied in the research community for decades, given their wide applicability. The goal of IASPs is to deduce the shape and/or constitutive properties of an object based on the acoustic scattering data due to an incident field collected at a set of receivers.  A diverse variety of application areas have this problem at their center, including sonar detection \cite{sonardetect}, nondestructive testing \cite{nondesttesting}, medical imaging \cite{sonartomog}, remote sensing \cite{remotesensing} and several more.

Inverse scattering problems can be addressed with both phase and phaseless data \cite{ammari2015phased}.
Methods utilizing phase data include the regularized Gauss-Newton method \cite{5161275}, recursive linearization methods \cite{reclin1, reclin2}, source inversion method \cite{van2021forward}, two-stage least squares method \cite{ITO2013211}, direct sampling methods \cite{Ito2012, Ito2013}. While being accurate, a downside is the difficulty of obtaining phase data in practical applications compared to phaseless data. Due to this fact, despite the phaseless reconstruction being significantly more ill-posed and non-linear, it is often preferred over phase-based reconstruction \cite{ning2024direct, Shuxin}. 

Several iterative methods using phaseless scattering data have been proposed to solve the inverse scattering problem \cite{Bao2016, Bao:13, Olha2007, Olha2010, 10.1121/10.0009851}. However, for iterative solvers, an intermediary shape is optimized by minimizing a loss function between its scattered field and the scattered field of the target shape. This process requires the execution of an expensive forward scattering solver at each optimization step, rendering the method impractical for several real-world use cases. Non-iterative methods such as sampling-based methods \cite{MargaretCheney2001, Ito2012, Potthast2006, ning2024direct} are faster, however they may not produce accurate results. These limitations underline the importance of developing more efficient and scalable methods to solve IASP.

In the last decade, machine learning and deep learning methods have been widely adopted in the scientific computing community as fast and data-driven alternatives to expensive iterative numerical solvers. Several deep learning methods have also been proposed to solve both the forward and the inverse acoustic scattering problems. In \cite{fan2020fast}, a convolutional neural network is used to learn a mapping between 2D obstacles and corresponding acoustic scattering far-field patterns. Later, in \cite{tang21acspnet} and \cite{meng-forward}, this idea is expanded to solve the forward acoustic scattering problem for 3D obstacles using a PointNet \cite{pointnet} encoder. For the inverse acoustic scattering problem, the proposed solutions are mainly focused on the 2D problem. A random forest model is used to perform surface shape reconstruction from phaseless acoustic scattering data in \cite{10.1121/10.0013506}. In \cite{WU2022108190, NAIR2023116167}, the authors propose physics-constrained neural network architectures to solve the acoustic inverse scattering problem for basic 2D shapes. In \cite{waqasahmed1} a pipeline of a forward and inverse networks are evaluated to reconstruct the 2D shapes of random scatterers from their  2D scattering cross-sections. In \cite{waqasahmed2}, the inverse design of an acoustic cloak is done by a forward and inverse neural network. In \cite{JUNQUEIRA2023116985}, the authors attempt to derive the interfacial defects on laminated surfaces by using a simple multi-layer perceptron. We refer the readers to \cite{XudongChen} for a comprehensive review of deep learning methods proposed to solve the inverse scattering problem. All these works highlight the potential and importance of the field, yet machine learning methods for solving inverse acoustic scattering problems for 3D shape reconstruction, to the best of our knowledge, remain unexplored.

In this paper, we propose ISSRNet (inverse scattering shape reconstruction network), a machine learning framework to solve the inverse acoustic scattering problem for retrieving the 3D shape of the scatterer, using phaseless acoustic scattering data from acoustically soft objects. We utilize scattering data obtained by illuminating the scatterer with a single incident wave (fixed angle, single frequency). 
The inversion framework consists of three different neural networks: a 3D shape auto-encoder, an inverse network, and a forward network. 
To optimize ISSRNet, we calculate a loss between the target and predicted shapes. This is in contrast to existing methods that calculate the optimization loss in terms of a derived quantity, viz., the difference between scattered field data from the target and predicted obstacles \cite{waqasahmed1, waqasahmed2, alsnayyan2022laplace}. As experiments will show, our approach performs very well.
Our main contributions of this work are as follows:
\begin{itemize}
    \item We propose a ISSRNet, a deep learning framework for 3D shape reconstruction from phaseless acoustic scattering data. As alluded to earlier, the approach we present relies on a different loss function, which is both a direct measure of performance and is more computationally different. 
    \item ISSRNet produces excellent results despite acting on limited scattering data, i.e., data obtained due to a single incident wave at a fixed frequency. This points to improvements that can be made with greater data diversity. 
    \item As will be shown, ISSRNet is evaluated on both the synthetic random particles and ShapeNet data sets. The reconstructions capture both the global properties of the scatterers as well as the local details and differentiate between different types of objects.
\end{itemize}
\section{\label{sec:2} Background and Motivation}
Consider an acoustically soft scatterer, $\Gamma$, embedded in a homogeneous medium $\Omega \in \mathbb{R}^3$. A time-harmonic ($e^{-i\omega t}$ dependence is assumed and suppressed) incident pressure wave with velocity potential, $\Phi^{i}(\vb{r})$, illuminates $\Gamma$, giving rise to a scattered wave with velocity potential, $\Phi^{s}(\vb{r})$. The resulting total velocity potential $\Phi^t (\vb{r}) = \Phi^{s}(\vb{r}) + \Phi^{i} (\vb{r}) $ satisfies the following boundary value problem,
\begin{subequations}
	\begin{align}
	\nabla \Phi^{t}(\mathbf{r}) + \kappa^{2}\Phi^{t}(\mathbf{r}) &= 0 \hspace{1cm} \textbf{r} \in \Omega     \label{eq:helmholtz_equation}, \\
	\Phi^{t}(\mathbf{r}) &= 0 \hspace{1cm} \textbf{r} \in \Gamma,     \label{eq:boundary_condition} \\
	\lim_{r\rightarrow \infty}\sqrt{r}\left (\frac{\partial \Phi^{s}}{\partial n} - i\kappa \Phi^{s} \right ) &= 0  \hspace{1cm} \textbf{r} \in \Omega,   
    \label{eq:rad_condition}
	\end{align}
	\label{eq:forward_problem}
\end{subequations}
where $\kappa$ is the wavenumber. Using an equivalence theorem, the scattering problem can be cast in terms of trace values of the velocity potential \cite{pierce2019,alsnayyan2020efficient}; we introduce the scattering cross-section (SCS) far-field operator $\mathcal{L}_{far}$ in terms of the surface pressure as 
\begin{equation}
\begin{aligned}
\mathcal{L}_{far}[\Phi,\Gamma] (\hat{\textbf{k}}) \doteq  \frac{1}{4\pi} \int_{\Gamma} \Phi(\mathbf{r^\prime}) e^{i\kappa\hat{\textbf{k}} \cdot \mathbf{r^\prime}}  d\mathbf{r}^\prime.
\end{aligned}
\label{eq:farfield}
\end{equation}
Here, the observation domain is on the unit $\hat{\vb{k}}$-sphere $S^{2}$, where $\hat{\vb{k}}(\theta,\phi) \in S^{2}$ is parametrized by $(\phi,\theta) \in [0,\pi] \times [0,2\pi]$. From hereon, this data is referred to as the scattered data or far-field data. The analytical solution of (\ref{eq:forward_problem}) is generally unavailable. The numerical solution of the integral equations is effected in a discrete setting using the boundary element method (BEM) \cite{alsnayyan2020efficient}. 

The goal of ISSRNet is to reconstruct the three-dimensional (3D) shape of the scatterer $\Gamma$, when scattering data on  $\hat{\vb{k}}(\theta,\phi)$ is available. In \cite{alsnayyan2022laplace} the authors develop a method for shape optimization problem, which relies on an iterative scheme, that perturbs an initial shape $\Gamma_{0}$, until its scattering data matches that of the target scatterer. While producing accurate reconstructions, this method requires the execution of the expensive forward scattering solver at each step of the optimization, which dominates the overall optimization process and becomes a serious computational bottleneck. To address this bottleneck, we formulate the inverse scattering problem for shape reconstruction using neural networks and define an objective function to optimize their parameters.

\subsection{Preliminaries: Neural network}

Following the generic formulation given in \cite{schmidt1992feed}, we mathematically define a neural network and its optimization procedure. A neural network can be represented by a function $f$, which maps an $n$-dimensional input feature space, to a $c$-dimensional latent space: $f: \mathbb{R}^{n} \rightarrow \mathbb{R}^{c}$. Neural network $f$ is parameterized by an $m$-dimensional weight vector $w \in \mathbb{R}^{m}$. Therefore we can express $f$ as $f(x, w)$, where $x \in \mathbb{R}^{n}$ is the input training data-point. Training the neural network involves updating the weights $w$, by minimizing the loss function $J: \mathbb{R}^{m} \rightarrow \mathbb{R}$. If we write the objective function $J$ in terms of network weights $w$, it takes the following form:

\begin{equation}
    \begin{aligned}
        J(w) = \frac{1}{n}\sum_{i}^{N}L(f(x^{(i)}, w), y^{(i)}) 
    \end{aligned}
\end{equation}

where $x^{i}$ and $y^{i}$ are the $i$-th input data-point and the corresponding ground-truth observation in the training data set $(x^{(i)}, y^{(i)})$ where $ 0<i<N$ and $N$ is the total number of training samples. The function $L$ is a term-wise loss function that measures the distance between the prediction made by the neural network and the ground-truth $y^{i}$. In our implementation, we use PyTorch \cite{autodiff}, which provides automatic differentiation for neural networks and a wide range of term-wise loss functions.

\subsection{Preliminaries: Shape encoding and mapping to scattered field data}

The proposed framework operates on a compressed latent shape representation space. This latent space is learned by the 3D shape auto-encoder, prior to the inverse mapping. The aim of the proposed framework is to map the input scattering data to this compressed shape latent space using an inverse neural network. For sake of simplicity, we only formulate the inverse neural network in this section and assume that the weights $w_{AAE}$ of the auto-encoder are learned \emph{a-priori}. The training process and the overall architecture of the auto-encoder are described in detail in Section \ref{subsub:autoenc}. Let $d(s, w_{AAE}) = pc$ be the pre-trained decoder of the shape auto-encoder that decodes shape latent vectors $s$ into their corresponding 3D point-clouds $pc$. We define an inverse neural network $inn(sc, w_{inn}) = s$, that aims to map input phaseless far-field scattering data $sc$ (formulated in \ref{eq:farfield}) to shape latent vectors $s$. Subsequently, we can define an end-to-end inverse scattering framework $d(inn(sc)) = pc$, that maps input scattering data $sc$, to 3D point-clouds $pc$, which represent the scatterers of interest. Given a training data set $(sc^{(i)}, pc^{(i)}), 0<i<N$, the objective of our training process is to find the set of values for the weights $w_{inn}$, such that $d(inn(sc^{(i)}), w_{AAE}) = pc^{(i)}_{G}$, where $pc^{(i)}_{G}$ and $sc^{(i)}$ are the $i$-th goal (ground-truth) point-cloud and input training scattering data respectively. This is equivalent to minimizing the following objective function:

\begin{equation}
\begin{aligned}
    J(w_{inn}) := \frac{1}{n} \sum_{i}^{N}CD( d(inn(sc^{(i)})), pc^{(i)}_{G})
\end{aligned} 
\end{equation}
\label{eq:obj_func_inv}

\begin{equation}
\begin{aligned}
    CD(P_{1}, P_{2}) = \sum_{x \in P_{1}} \min_{y \in P_{2}} ||x-y||_{2}^{2} + \sum_{y \in P_{2}} \min_{x \in P_{1}} ||x-y||_{2}^{2}
    \label{eq:chamfer}
\end{aligned}
\end{equation}

The term-wise loss function to optimize the inverse network is Chamfer distance (CD), defined in Equation \ref{eq:chamfer}. As it can be seen from the formulation, $CD$ is a metric that quantifies the distance between two point clouds by summing the distances between each point and its closest neighbor in the other cloud.

\section{\label{sec:3} ISSRNet}

Next, we present the different modules and methods that compose our predictive framework. These include data generation/preprocessing and neural network components. In our experiments, we evaluate the proposed method on both the synthetic random 3D particle dataset as well as the widely ShapeNet \cite{shapenet}, a used 3D computer vision benchmark dataset.

\subsection{Pre-processing and Generation of Geometry Data}
\label{sec:data_gen}

\begin{figure*}[hbt!]
    \centering
    \includegraphics[width=\textwidth]{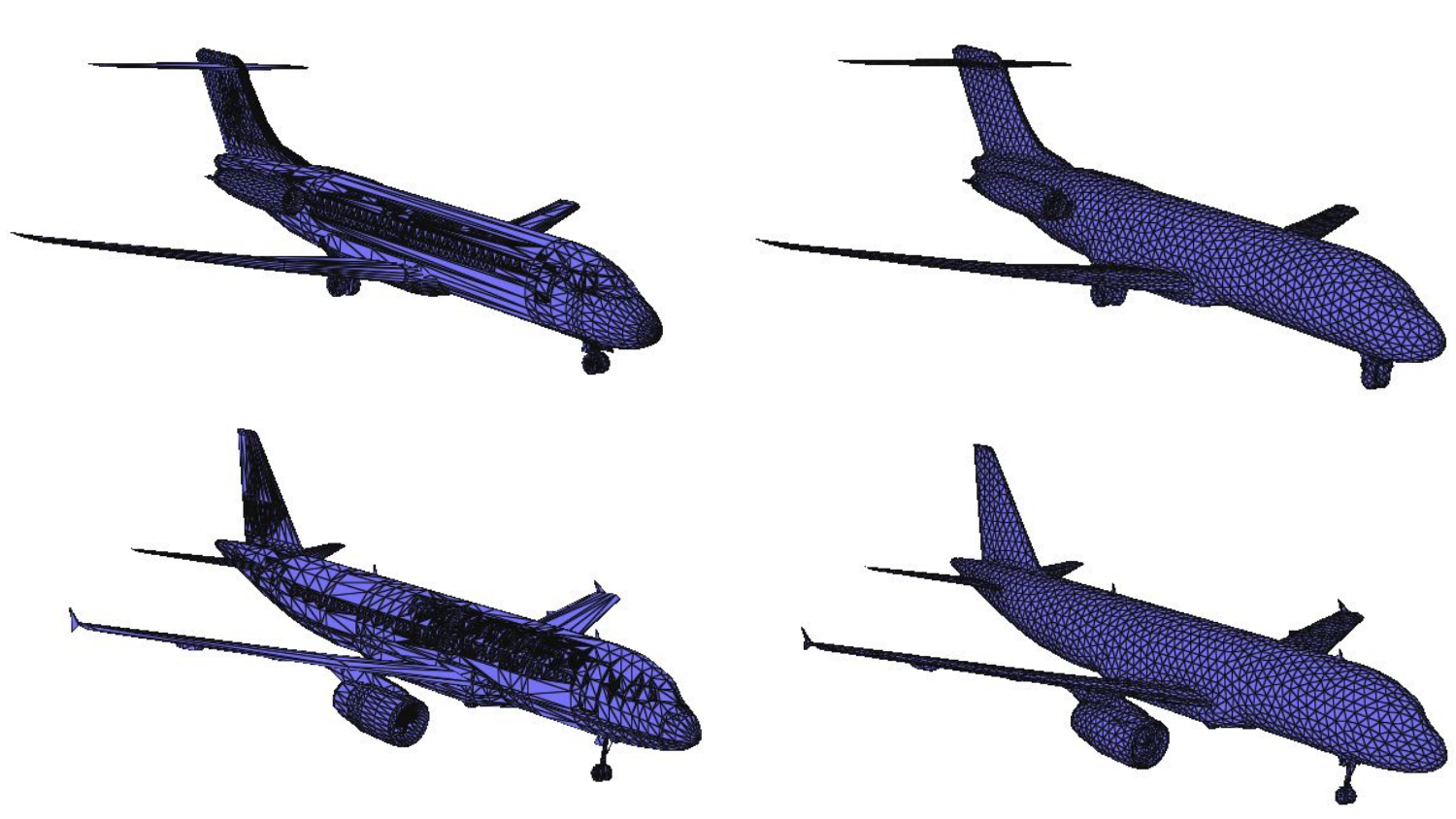}
    \caption{Data pre-processing step for ShapeNet meshes. The left column contains the original meshes that are not watertight and have poor-quality triangulation. The meshes in the right column show the remeshing product.}
    \label{fig:Figure1}
\end{figure*}
\begin{figure}[hbt!]
    \centering
    \includegraphics[width=0.5\textwidth]{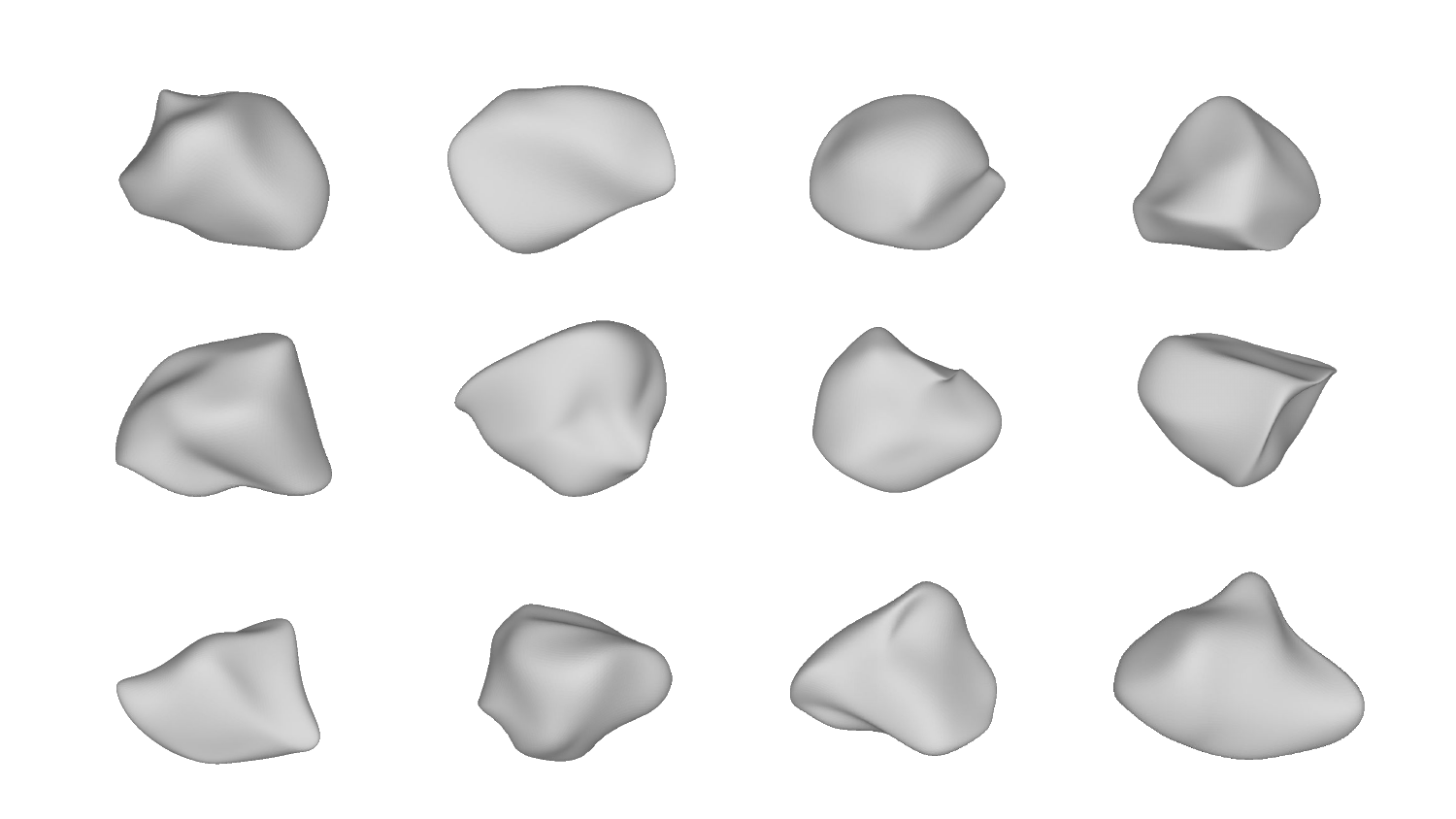}
    \caption{Random particles from the data set}
    \label{fig:Figure2}
\end{figure}

In this subsection, we explain the data generation and pre-processing steps utilized. Training the proposed networks requires the computation of scattered fields for all shapes in the training data set. We use the solver introduced in \cite{alsnayyan2020efficient} to compute the scattering far-fields. As with any physics based solver, it requires high quality tesselation (sufficiently fine to capture the underlying physics,  conformal elements, elements with the right aspect ratio, watertight, etc). This is a challenge for available meshes that are intended for visualization and not for computational physics. 

In order to overcome this practical issue, we utilize two remeshing methods to pre-process our shape data. The first step is to make the scatterer meshes watertight. We utilize ManifoldPlus \cite{manifoldplus}, which is a scalable and robust tool developed to generate watertight surface meshes from triangle soups. After the mesh is transformed into a watertight mesh, we use geogram, which utilizes anisotropic smooth remeshing methods presented in \cite{anistropic1, anisotropic2}. The original mesh in ShapeNet and the watertight remeshed version is shown Figure \ref{fig:Figure1}. In this 2-step pre-processing phase, the number of triangles in the final re-mesh is also configurable, therefore this provides an easy way of adjusting the average edge length in our scatterer meshes, a necessary feature to accurately capture the physics.

We use two sets of data to train the network; one with random particles and the other with real geometries. These are described next.

\subsubsection{Random 3D Particle Data Generation}
The random particle data  generation process consists of random 3D shape generation and the corresponding scattered field computation. To generate the random 3D shapes, the random particle generator introduced in \cite{particlegen} is used.

The particle generator utilizes low-frequency spherical harmonics to determine shape properties such as elongation, roundness and aspect ratio, based on the shape analysis performed in \cite{zhao2017}. The process yields a variety of random particles which can have sharp, non-convex and flat features. These shape properties introduce complex patterns in the resulting scattered fields, increasing variation. In addition, the shape generator uses an evenly subdivided icosahedron mesh for each particle as a starting point, therefore the data set does not have any mesh quality problems. The idea is similar to the data generation method used in \cite{fan2020fast}, however the 3D shapes in this work can have variations in all 3D directions and are not limited to convex prisms. Figure \ref{fig:Figure2} shows samples from the random particle data set.
\begin{figure*}[hbt!]
    \centering
    \includegraphics[width=\textwidth]{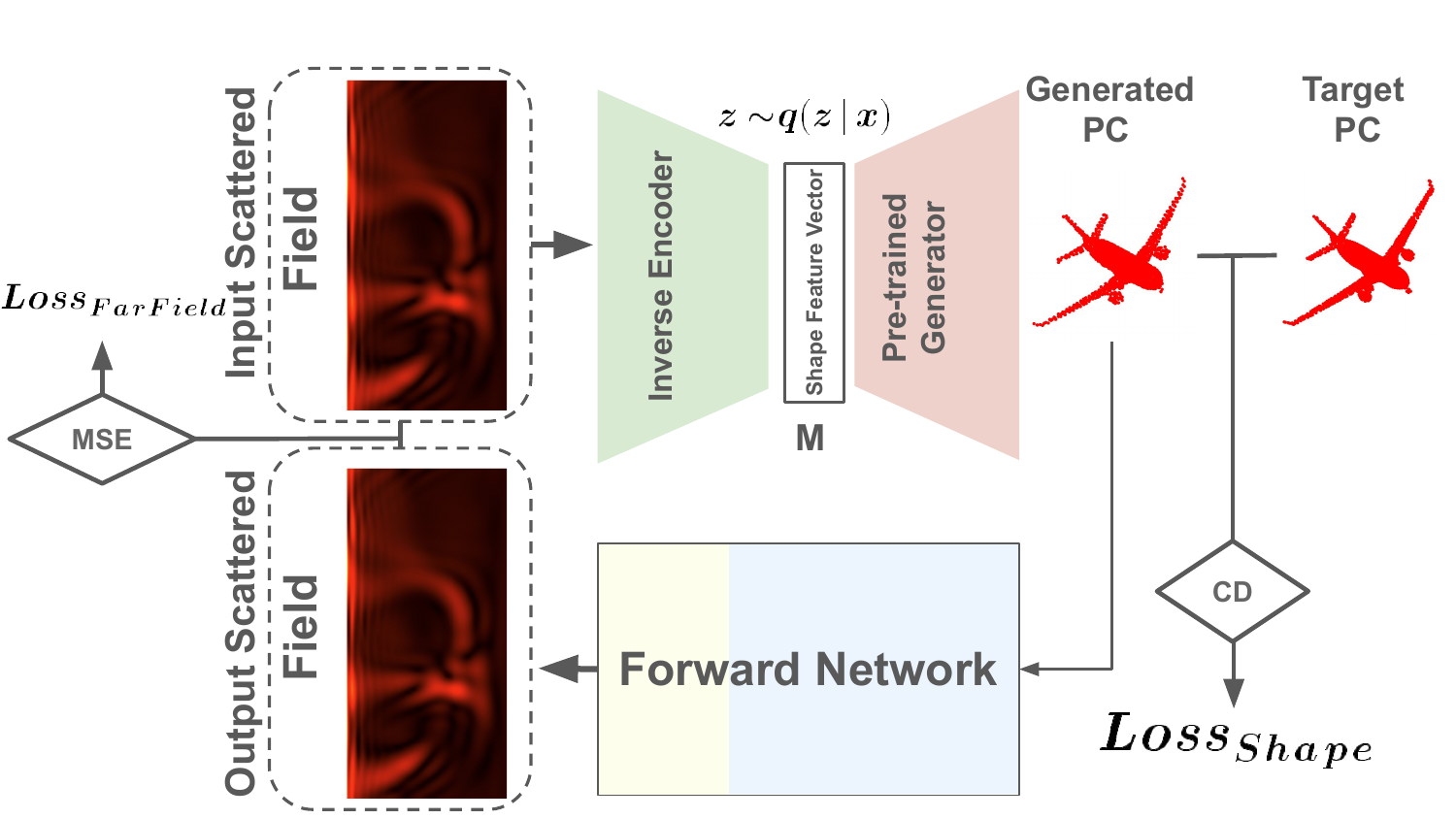}
    \caption{Proposed deep learning pipeline. The color-coded pre-trained generator and inverse encoder modules can be found in Figures \ref{fig:Figure4} and \ref{fig:Figure5}, respectively. The pipeline can be trained using two different loss functions, $Loss_{Shape}$ which is the Chamfer distance between the generated and target point-clouds; $Loss_{FarField}$ which is the mean-squared error between the output and input scattered fields. }
    \label{fig:Figure3}
\end{figure*}

\subsubsection{ShapeNet Pre-processing}
\label{sec:snet_preprocess}
Popular 3D shape datasets such as ShapeNet or ModelNet \cite{modelnet, shapenet} include a rich variety of meshes belonging to different classes of objects. We use two classes in our analysis here. As was alluded to earlier, these meshes are not made for analysis and have to be modified using the procedure described earlier.

\subsection{Neural Network Modules}
\label{sec:nnet_modules}

The predictive end of the proposed framework consists of a pipeline of three different neural architectures: A 3D variational auto-encoder, a convolutional inverse network, and a forward network. Let \textit{$\mathcal{PC}$} be the set of 3D scatterer point-clouds and \textit{$\mathcal{SC}$} the corresponding set of 3D acoustic scattered far-fields. The auto-encoder, the inverse network and the forward network are trained to learn the mappings $\textit{$\mathcal{PC}$} \rightarrow \textit{$\mathcal{PC}$}$, $\textit{$\mathcal{SC}$} \rightarrow (\textit{$\mathcal{PC}$} \rightarrow \textit{$\mathcal{PC}$})$ and $\textit{$\mathcal{PC}$} \rightarrow \textit{$\mathcal{SC}$}$, respectively. Note that the inverse mapping is not directly between \textit{$\mathcal{SC}$} and \textit{$\mathcal{PC}$}. The inverse network instead learns a mapping from \textit{$\mathcal{SC}$} to the 3D shape latent space, $\textit{$\mathcal{PC}$} \rightarrow \textit{$\mathcal{PC}$}$, which is learned by the 3D auto-encoder. 

Figure \ref{fig:Figure3} shows the overall deep learning framework. The goal of the framework is to learn the mapping $\mathcal{SC} \rightarrow \mathcal{PC}$. To this end, first the 3D shape latent space $\mathcal{PC} \rightarrow \mathcal{PC}$ is learned by the auto-encoder. Then, the inverse network learns a mapping from $SC$ to this latent space. Each M-dimensional vector from the latent space represent a 3D point-cloud. Since these vectors are samples from $\mathcal{PC} \rightarrow \mathcal{PC}$, they can be decoded by the auto-encoder into 3D point-clouds. After the intermediary (predicted) scatterer shape is produced by the pre-trained generator (red section in Figure \ref{fig:Figure3}), there are two approaches we consider, to calculate a loss function to optimize the inverse encoder. The first approach,(see \ref{eq:obj_func_inv} in Section \ref{sec:2}), which makes the training process completely independent from the forward solution, operates the loss solely on the target (3D shape) space by employing a Chamfer distance (see \ref{eq:chamfer}) between the predicted and target point-clouds of the scatterers. The second approach calculates a loss on the input-space, to indirectly morph the intermediary point-clouds. This can be achieved by feeding the generated point-cloud into the pre-trained forward network ($PC \rightarrow SC$) to predict the scattered far-field information. Since the shape is optimized based on the loss between target and intermediary scattered fields, this method is similar to the existing iterative and 2D ML methods  \cite{alsnayyan2022laplace, waqasahmed1, waqasahmed2}. Therefore, we implement and compare both approaches to investigate the necessity and/or improvement effects of utilizing the second approach. The two approaches are also visualized in Figure \ref{fig:Figure3}. We experiment with optimizing the network using only $Loss_{Shape}$ and with adding $Loss_{FarField}$ as a regularizing term. As it can be seen from the figure, calculating $Loss_{FarField}$, requires the extra step of generating a scattered field from the generated point-cloud, using the forward network.

\subsubsection{3D Variational Auto-encoder}
\label{subsub:autoenc}
The proposed predictive framework operates on a configurable, smooth latent space, representing each 3D scatterer in the data set. This compressed vector representation provides an advantage when designing the inverse network, since the target output becomes a single size-configurable vector. This allows us to easily experiment with very compact representations for the 3D scatterers. To learn the latent space from 3D point clouds, we adopt the variational auto-encoder architecture proposed in \cite{3daae}. Figure \ref{fig:Figure4} shows the network architecture.
\begin{figure}[hbt!]
    \centering
    \includegraphics[width=0.5\textwidth]{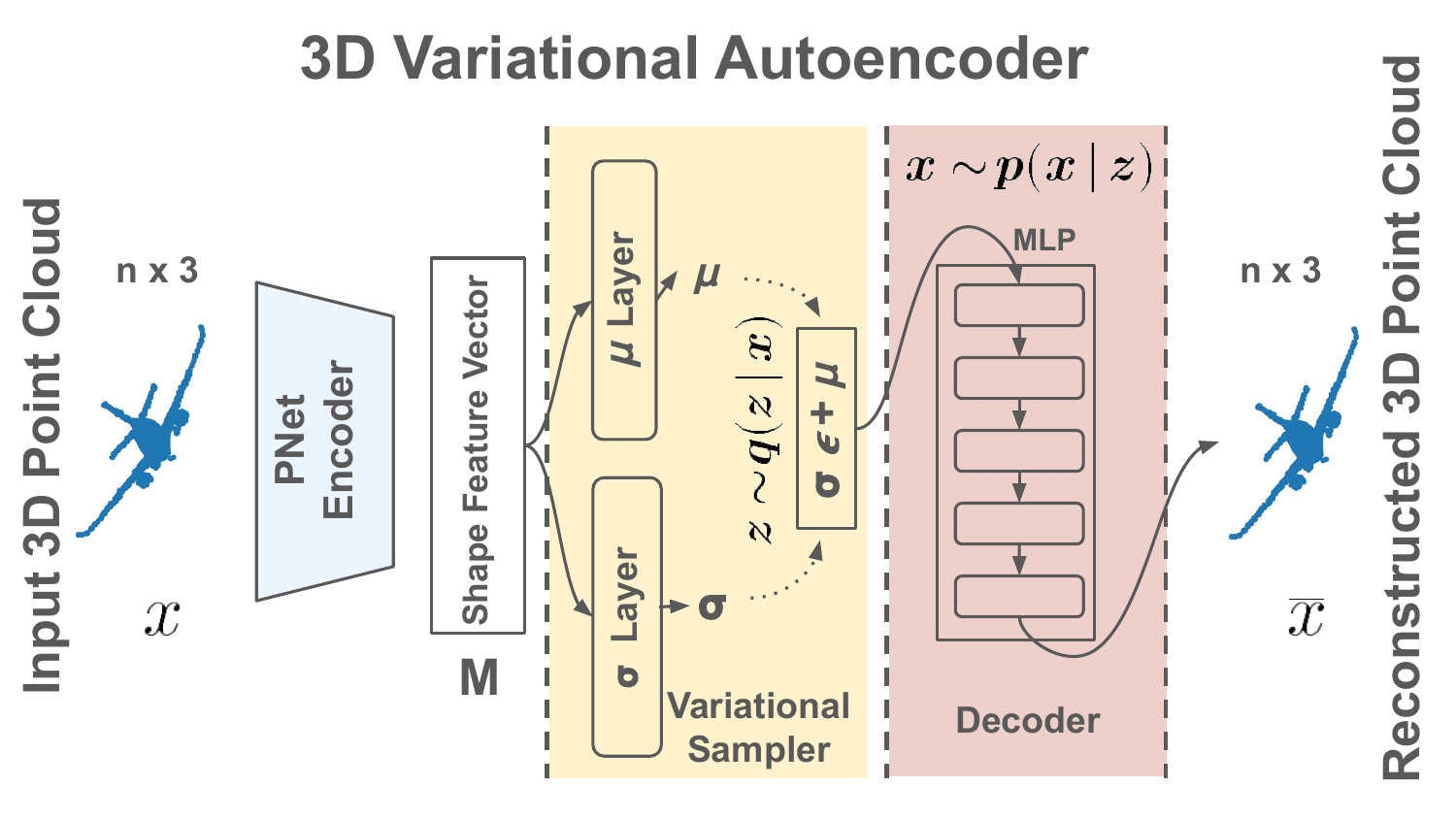}
    \caption{(color online) 3D Auto-encoder Architecture}
    \label{fig:Figure4}
\end{figure}

The input 3D point-cloud $x$ of the scatterer of interest is first fed into the PointNet Encoder, shown in Figure \ref{fig:Figure6}. This component encodes the input point-cloud to an M-dimensional global feature vector. The next step is to learn a mapping from this feature vector, back to the original point-cloud. The architecture utilizes a variational auto-encoding approach to achieve this goal. The goal in variational auto-encoders is to learn an approximation $q(z|x)$ to the posterior distribution $p(z|x)$ for the training data set $X$, where data points $x \in X$, when a known prior distribution such as the normal distribution $p(z) = \mathcal{N}(\mu, \sigma^{2})$ is given. Therefore, given a data point $x$, the process can generate the code $z \sim q(z|x)$, which approximates $p(z|x)=\mathcal{N}(\mu, \sigma^{2})$, the probability distribution over all possible values of the input $x$. This approach allows the model to learn a generative latent space for the scatterers, where samples from it are similar to the training data, and the statistical properties of the underlying distribution are interpretable, thanks to the approximation to the prior normal distribution. To draw a random sample from the learned latent $z$, the model utilizes a technique known as the ``reparametrization trick". The sample $z \sim q(z|x)$, where $p(z|x)=\mathcal{N}(\mu, \sigma^{2})$, can be reconstructed as $z= \sigma\epsilon + \mu$, where $\epsilon \sim \mathcal{N}(0, 1)$. This trick allows backpropagation to work with the random sampling involved, since the sampling is performed via the deterministic function $z= \sigma\epsilon + \mu$. The final step is to approximate $x \sim p(x|z)$ with the generator multi-layer perceptron (MLP), which given a code $z$, reconstructs $x$ as $\overline{x}$. We refer the readers to \cite{varaes} for further details and explanation.

\begin{equation}
\begin{aligned}
    KLD(P||Q) = \sum_{x \in X} P(x)log\frac{P(x)}{Q(x)}
    \label{eq:kld}
\end{aligned}
\end{equation}

\begin{equation}
\begin{aligned}
    Loss_{VAE} = CD(P_{1}, P_{2}) + KLD(p(z)||q(z|x))
    \label{eq:loss_comb_vae}
\end{aligned}
\end{equation}

The weights $w_{VAE}$ variational auto-encoder (VAE) are trained using two term-wise loss functions: the shape reconstruction loss and the variational loss. The former measures the 3D Chamfer distance (CD) (Equation \ref{eq:chamfer}) between the input and output point-clouds $x$ and $\overline{x}$, while the latter is the Kullback-Leibler divergence (KLD)(Equation \ref{eq:kld}) between the prior distribution $p(z)$ and latent $q(z|x)$. The combined loss function then becomes $Loss_{VAE}$ (Equation \ref{eq:loss_comb_vae}).

\begin{figure*}[hbt!]
    \centering
    \includegraphics[width=\textwidth]{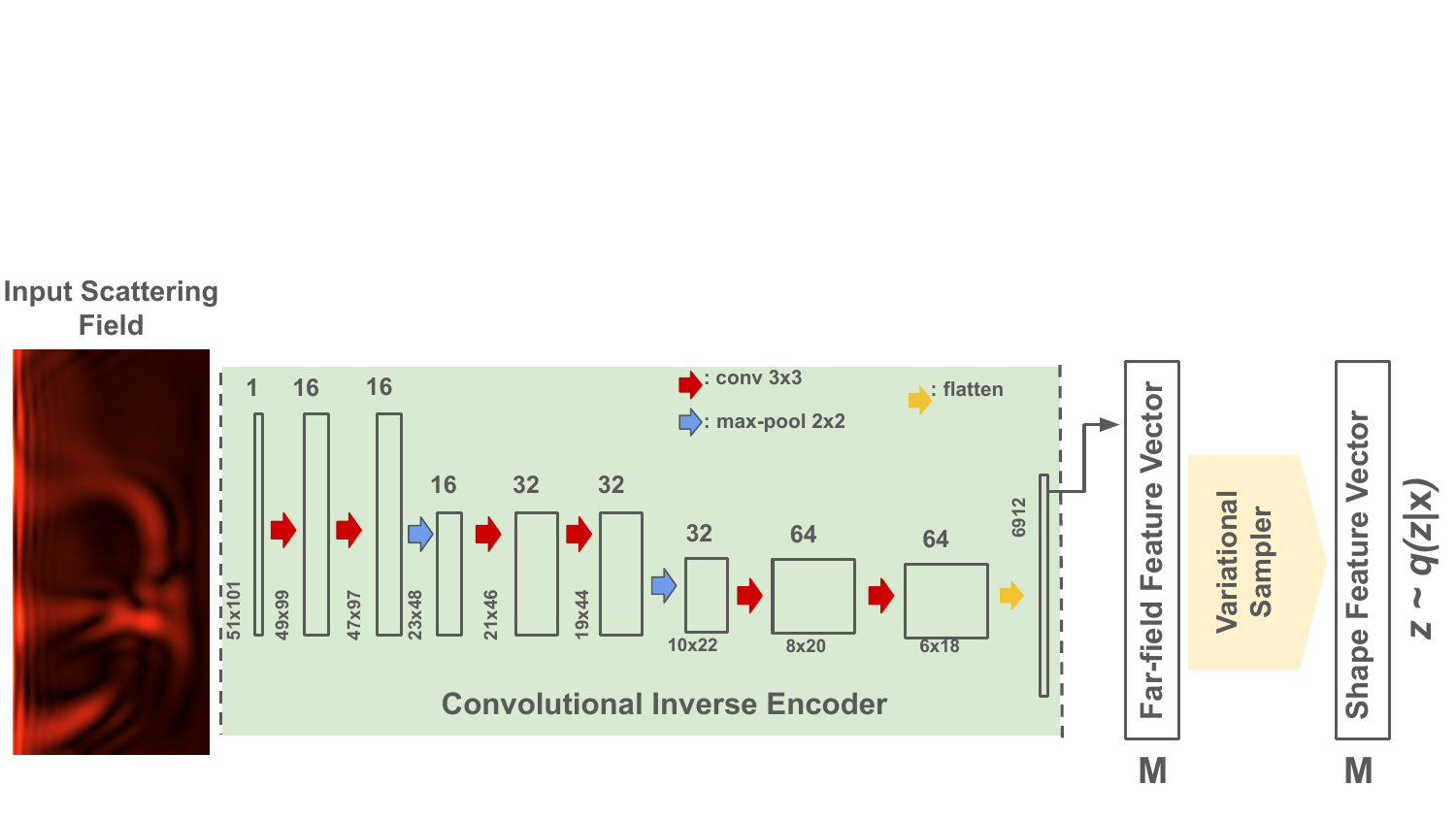}
    \caption{Inverse network architecture. A convolutional neural network is used to perform feature extraction from the input scattered field. The resulting shape feature vector is then processed with the variational sampler to sample from $q(z|x)$, which can be decoded using the pre-trained auto-encoder generator. }
    \label{fig:Figure5}
\end{figure*}

\subsubsection{Inverse Network}
The inverse network in the predictive pipeline learns a mapping between the scattered fields (input space) and latent vectors that are compatible with the pretrained 3D variational auto-encoder generator $x \sim p(x|z)$, shown in \ref{fig:Figure4} (red section). The latent space of the auto-encoder allows the inverse encoder to operate on a smooth optimization surface, similar to a Gaussian distribution. The encoded feature vector is decoded by the pre-trained generator to produce the 3D point cloud, representing the scatterer.

We extract local features from the input scattered field using a 2D convolutional neural network. The convolutional encoder is shown in Figure \ref{fig:Figure5}. The encoder outputs an M-dimensional shape feature vector, which is fed into a variational sampler module (see yellow portion in Figure \ref{fig:Figure4}), to sample a random instance from the latent $q(z|x)$. The inverse network is updated with the loss function $Loss_{Inverse} = Loss_{Shape} + \alpha_{FF} Loss_{FarField}$. Here, $\alpha_{FF}$ is a factor that controls how much of the forward loss we want to incorporate into $Loss_{Inverse}$. In our experiments, we use different values for $\alpha_{FF}$ to observe potential improvements (see Section \ref{sec:4}). 

\subsubsection{Forward Network}
As explained in the beginning of Section \ref{sec:nnet_modules} and shown in Figure \ref{fig:Figure3}, the pipeline can utilize two different loss functions to optimize the parameters of the inverse encoder. The calculation of $Loss_{FarField}$ requires the computation of the scattered field from the generated intermediary point-cloud. Since a numerical solution such as the BEM solver is too expensive to employ in such a training scenario, a forward neural network is instead trained and utilized.

The forward network is trained to learn a mapping from 3D shape properties to the acoustic scattering information of the obstacles of interest. This problem is known as the forward acoustic scattering problem, formulated in Equation \ref{eq:forward_problem}. In \cite{meng-forward, tang21acspnet}, the authors propose the first 3D deep learning framework to solve the forward problem for arbitrary 3D obstacles. They utilize the popular PointNet architecture to embed the obstacle point-clouds into a global feature vector. Then a fully-connected decoder maps these feature vectors to the spherical harmonics coefficients of the corresponding scattered field. We adopt a similar approach; however, we directly use the scattered far-fields as our output, instead of spherical harmonic coefficients.

\begin{figure}[htb]
    \centering
    \includegraphics[width=0.5\textwidth]{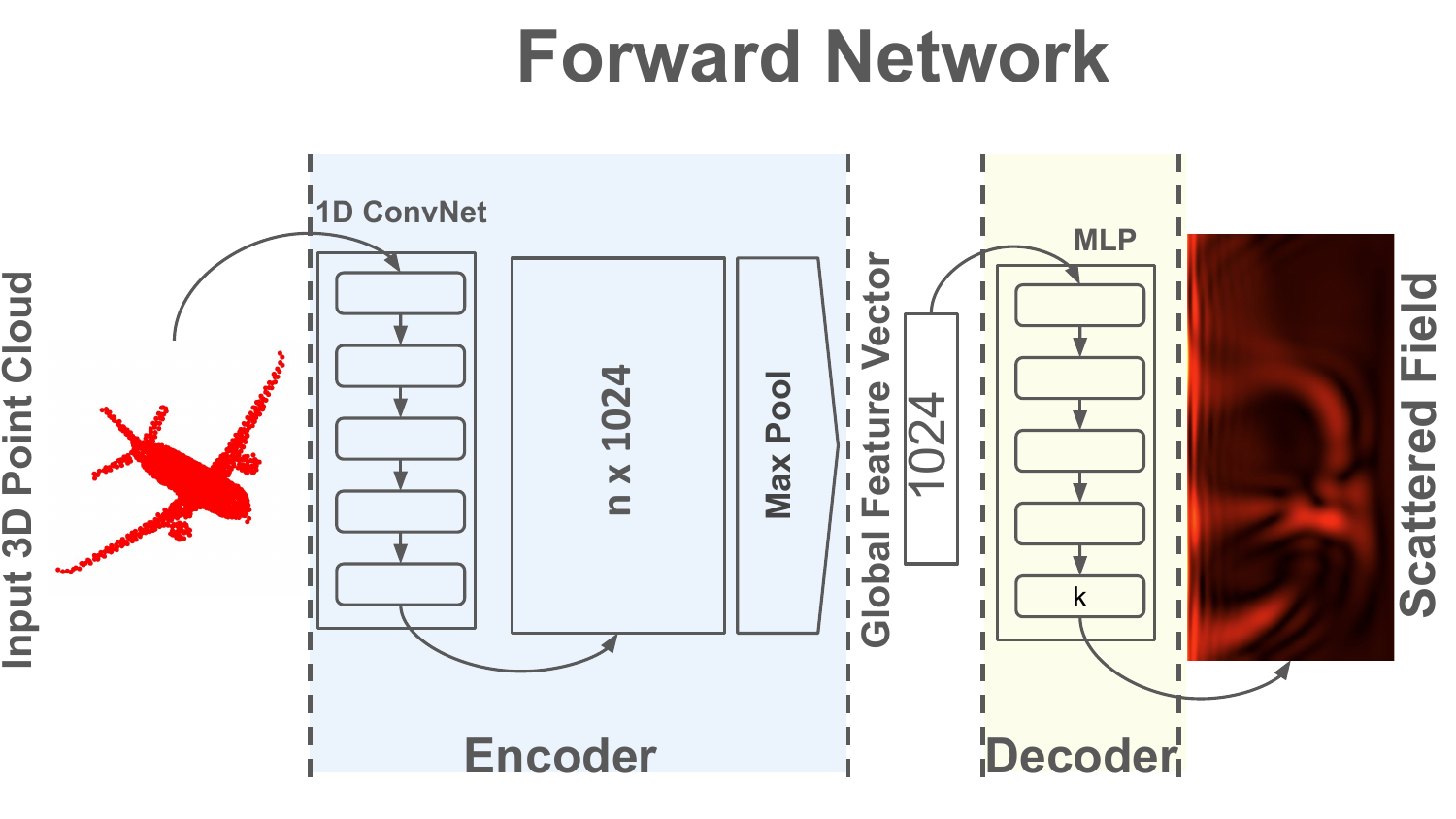}
    \caption{Forward network architecture. The 3D features are first extracted from the input point cloud, with the PointNet encoder. Then, the feature vector is mapped to the scattered field through a multi-layer perceptron. }
    \label{fig:Figure6}
\end{figure}

A 3D point-cloud is first fed into a 1D convolutional encoder. This encoder expands each 3-dimensional Cartesian point into 1024-dimensional latent points. Then, the global max-pooling layer reduces the global feature matrix into a global feature vector. This 1024-dimensional vector stores only the maximum for each feature, i.e. stores only the information relevant to the most important points. The global feature vector is finally input into the MLP decoder, which maps the feature vector to the corresponding scattered field. Figure \ref{fig:Figure6} shows the forward network architecture.

\begin{figure*}[!htbp]
    \centering
    \includegraphics[width=\textwidth]{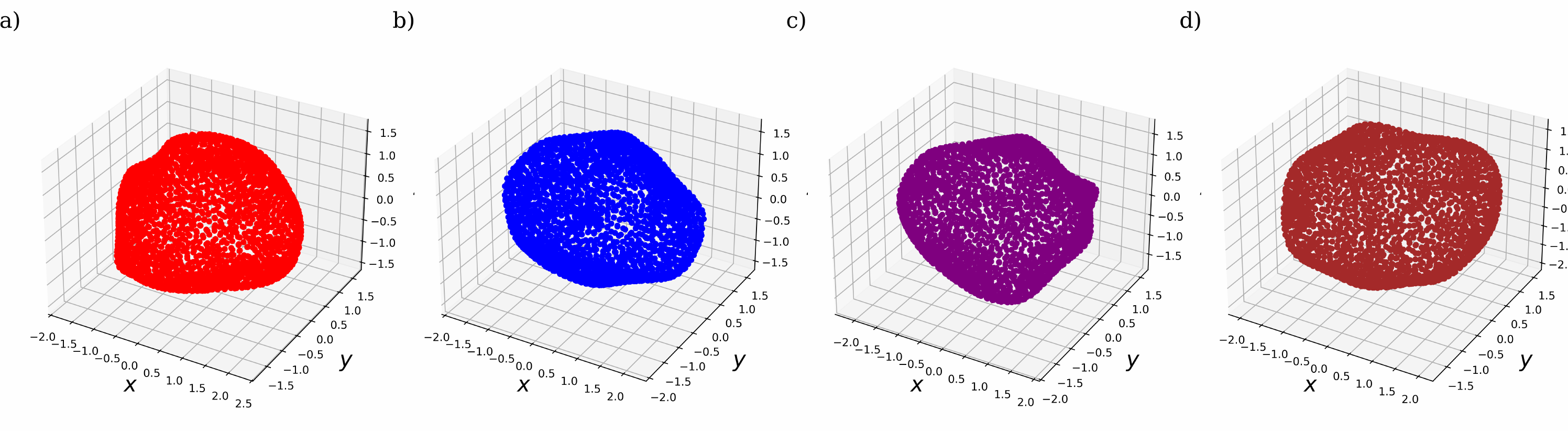}
    \includegraphics[width=\textwidth]{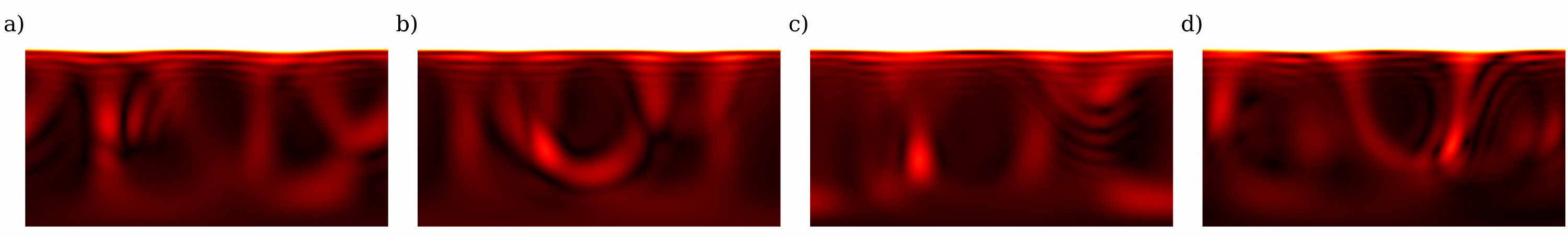}
    \includegraphics[width=\textwidth]{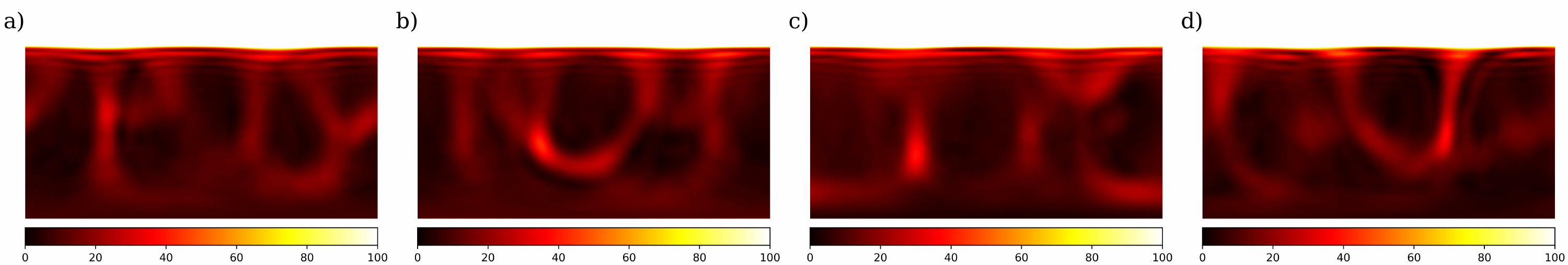}
    \caption{Reconstruction results for the Forward Network. The first row shows the point-clouds of interest, the second row shows the ground-truth scattered fields computed by the numerical solver, and the third row shows the scattered fields predicted by the forward network. The average relative L2 error is $0.05$ for the test set of 5000 particles.}
    \label{fig:Figure7}
\end{figure*}
\begin{figure}[!htbp]
    \centering
    \includegraphics[width=0.5\textwidth]{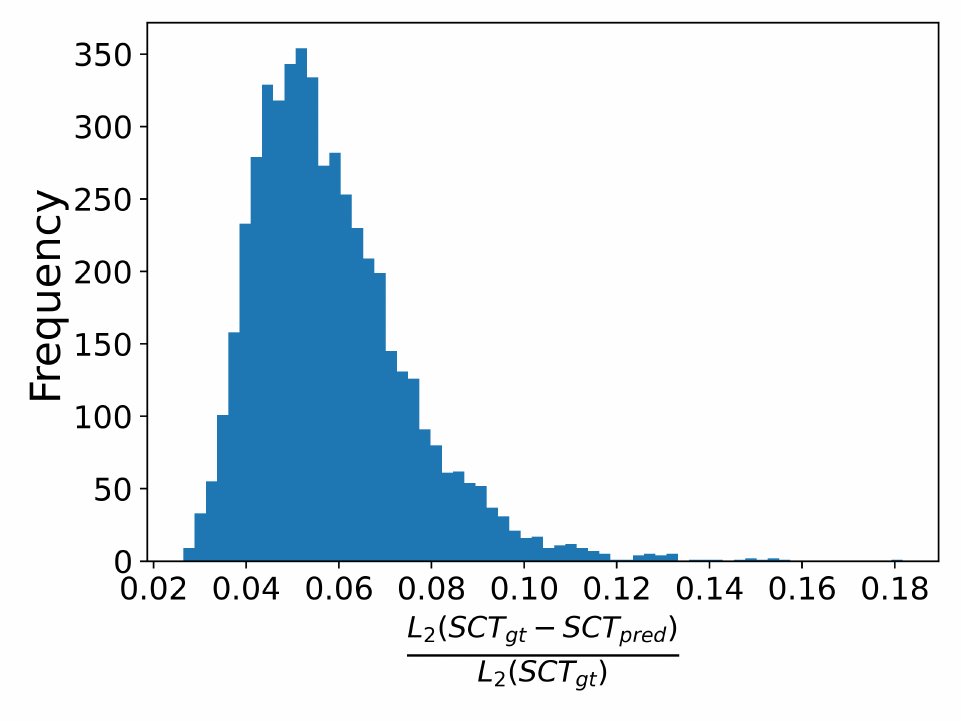}
    \caption{Error distribution histogram of the reconstructed random particle scattered fields by the forward network.}
    \label{fig:Figure8}
\end{figure}

\begin{figure*}[!htbp]
\centering  \includegraphics[width=0.9\textwidth]{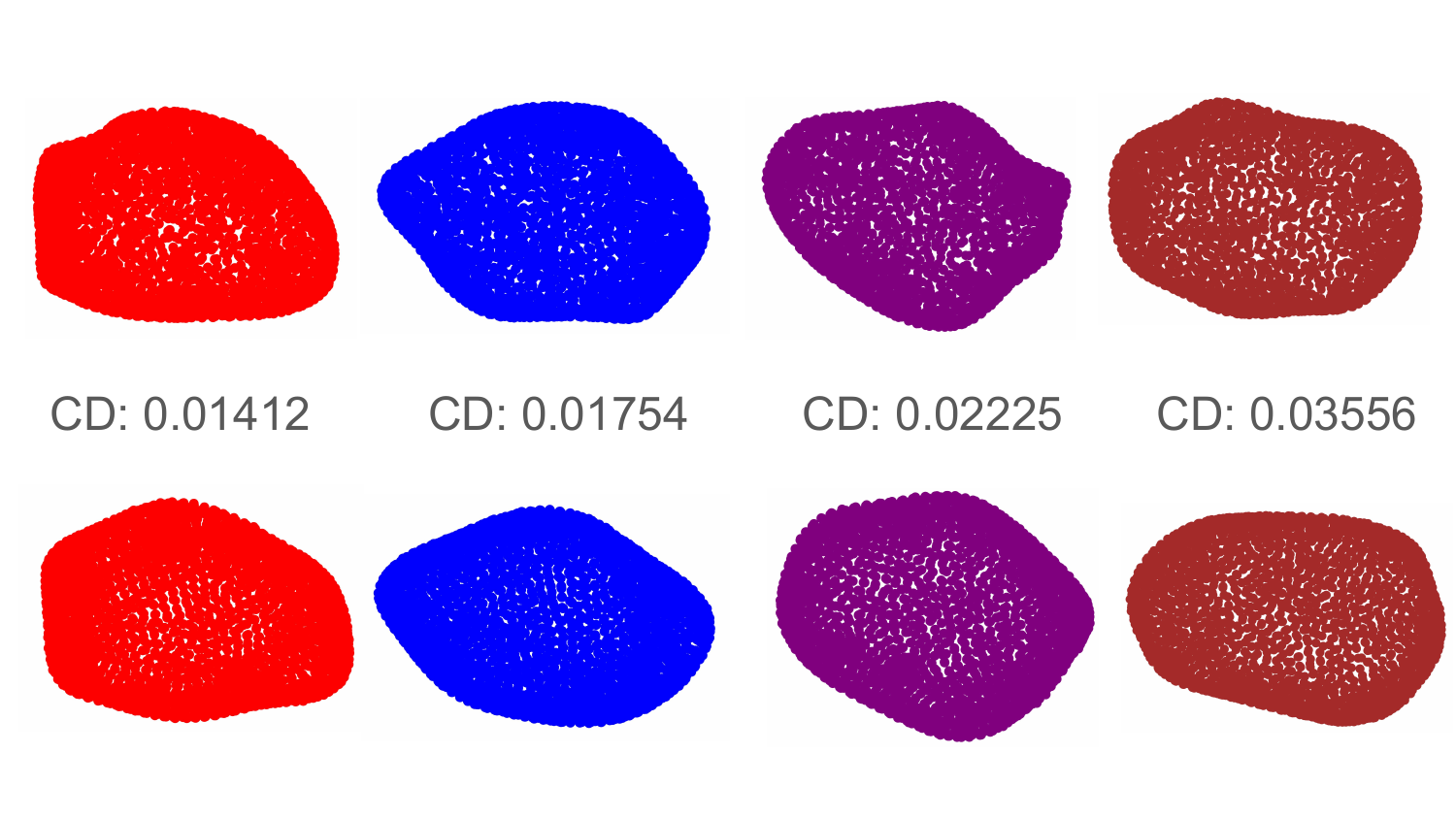}
\caption{Ground-truth (top) and reconstructed (bottom) scatterer point-clouds of test samples from the random particles data set. The reconstruction error in Chamfer distance (CD) is given for each sample.}\label{fig:Figure9}
\centering
\end{figure*}
\begin{figure}

    \includegraphics[width=0.5\textwidth]{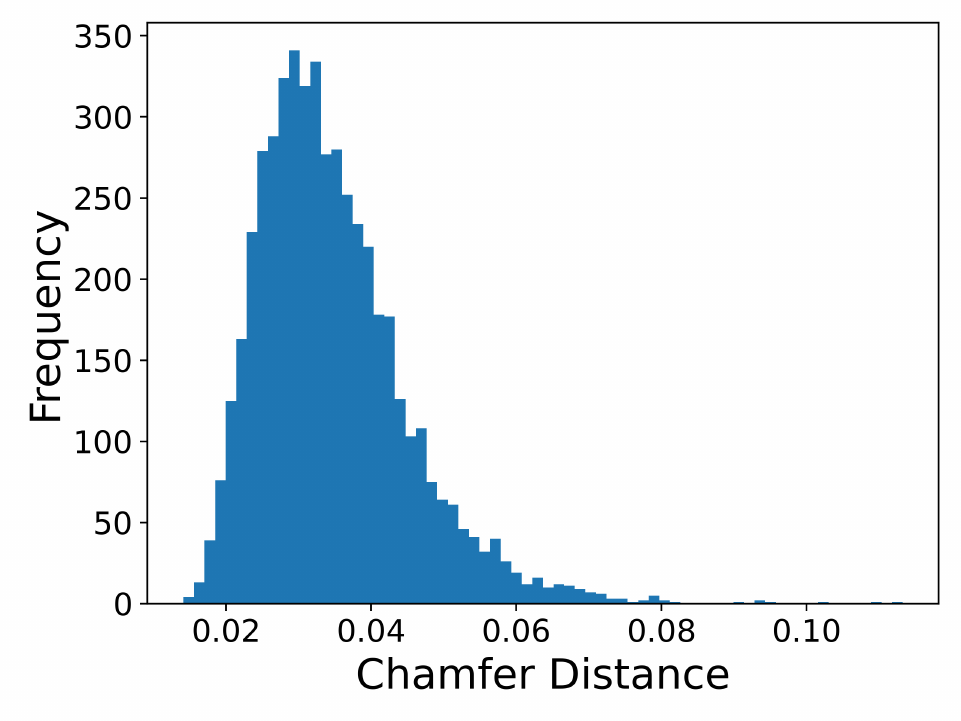}
    \caption{Error distribution of shape reconstruction results for 5000 test samples from the random particles data set }
    
    \label{fig:Figure10}

\end{figure}

\section{\label{sec:4} Results and Discussion}

In this section, we present the 3D reconstruction results obtained by the proposed framework. 

\subsection{Experimental Setup}
For our experiments, we consider two cases. 
First, we evaluate the proposed method on the random smooth particle data set. We randomly generate 50000 random particles with the method described in Section \ref{sec:data_gen}, then compute the scattered fields at 600 Hz, using the BEM solver. The fields are computed at 51 latitudinal and 101 longitudinal Gauss-Legendre quadrature coordinates.
Next, we evaluate the proposed method on the \textit{airplane} and \textit{cars} classes of the popular 3D vision benchmark data set ShapeNet \cite{shapenet}. We first process the data set with the preprocessing step explained in Section \ref{sec:snet_preprocess}. Then, we calculate the scattered fields at 750 Hz at the same Gauss-Legendre points as in the random particle data set. The \textit{cars} contain 3146 samples and \textit{airplanes} classes consists of  3227 sample. Objects from both classes are normalized into a bounding sphere with approximately $\textit{r}=2m$. The point clouds representing the scatterer meshes are sampled using the furthest point sampling algorithm, and the sample size is $2048$. For the embedding size, we select $M=64$ (see Section \ref{sec:3}). 
All neural network models use the LeakyReLU activation function, with the negative slope parameter set to $10^{-2}$. We optimize all models using the Adam optimizer with weight decay hyperparameter set to $10^{-4}$. In order to schedule the learning rate, we use a cosine annealing learning rate scheduler \cite{loshchilov2017sgdr} and set the initial learning rate to $5e-4$. For each data set, we use a training-testing split ratio of $9:1$. 

All experiments are run on a single node equipped with an Intel Xeon 8358 CPU with 256GB of memory and a single NVIDIA A100-40GB GPU.

\subsection{\label{subsec41} Case 1: Random Particles}

In this section, we discuss the evaluation results of the proposed framework, on the random particle data. This data set contains globally round and smooth objects with random local perturbations. These perturbations can result in sharp, non-convex and/or flat local features, which can have very distinct scattering properties. At a first glance, the variation in the random particle data set might look minor. However, indirectly differentiating the subtle local differences between shapes that globally agree is a challenging task in the context of a learning problem. We consider this step as a warm-up and tuning step for our experiments with ShapeNet data. In this experiment, we train the framework using both $Loss_{Shape}$ and $Loss_{FarField}$, as explained in Section \ref{sec:3}. 

We first start by evaluating the forward network, trained with the random particle point-clouds and the corresponding scattered fields. Figure \ref{fig:Figure7} shows the field reconstruction results for random particles drawn from the test set of 5000 samples. As it can be observed from the results, the forward network is able to capture the global structure of the scattered fields. However, local details are sometimes mispredicted and/or smoothened by the forward network. Figure \ref{fig:Figure8} shows the error distribution for the test samples, evaluated by the forward network. For measuring the reconstruction error for the scattered fields, we use the relative L2-norm of the difference between the ground-truth scattered field $SCT_{gt}$ and the predicted scattered field $SCT_{pred}$, so $Relative_{L_{2}} = \frac{L_{2}(SCT_{gt}-SCT_{pred})}{L_{2}(SCT_{gt})}$. As it can be observed from the error distribution histogram, most reconstruction errors are accumulated around 5\%.

\begin{table}[!htbp]
\centering
\caption{Test loss of the proposed models on the synthetic random particle (SYNT), ShapeNet airplane (AP) and ShapeNet cars (CARS) data sets. $\alpha_{FF}$ is the factor determining how much of the forward (far-field) loss we add to the optimization loss.}
\begin{tabular}{*7c}
\toprule
$\alpha_{FF}$ &  \multicolumn{1}{c}{SYNT} & \multicolumn{1}{c}{AP} & \multicolumn{1}{c}{CARS}\\
\midrule

 0.0  &  \textbf{0.034}  & \textbf{0.017}  & \textbf{0.0093} \\
 0.25 &  0.034  & 0.018  & 0.0094 \\
 0.50 &  0.034  & 0.018  & 0.0094 \\
 0.75 &  0.034  & 0.018  & 0.0095 \\
\bottomrule
\label{table:table_loss}
\end{tabular}
\end{table}

After verifying the reconstruction capability of the forward network, we continue with our experiment by training the shape auto-encoder and the inverse network, using the random particle data set. We aim to determine the effect of utilizing the forward pass to optimize the model. In order to do this, we use the loss function defined as $Loss_{Inverse} = Loss_{Shape} + \alpha_{FF}\times Loss_{FarField}$ (see Section \ref{sec:3}). Here $\alpha_{FF}$ is a tunable hyperparameter that controls the amount of $Loss_{FarField}$ we want to include for the training procedure. Table \ref{table:table_loss}, shows the loss values for different $\alpha_{FF}$ values of $0.0$, $0.25$, $0.50$ and $0.75$, as higher factors did not result in any improvements. As it can be seen from the loss values, using a composite loss of both shape and scattering data does not improve the results. Figure \ref{fig:Figure9}, shows the reconstruction results for the random particle data set, with the forward step bypassed by setting $\alpha_{FF} = 0$. We can see that the proposed framework is able to capture the global structure. However, we see a significant smoothing of sharp features. Note that we don't observe such degree of smoothing in the ShapeNet reconstructions, which are presented in the next subsection. Moreover, the Chamfer distance error distribution for $5000$ reconstructed test shapes for the random particles data set shows that most reconstruction errors are accumulated around $0.03$, which is a higher error average than both ShapeNet results (see Table \ref{table:table_loss}. This suggests that, despite ShapeNet dataset containing more complex structures, the random particles data set provides a more challenging learning task for the framework. This is due to the fact that the global structure of airplanes (the position of the wings, body, tail etc.) and that of cars (the position of the wheels, body, windshield etc.) are much more well-determined through-out the data set. This results in more predictable shape perturbations for different training samples. On the other hand, the random particles share the global spherical structure, but the local perturbations are much more unpredictable, increasing the random variation. Subsequently, this makes it more difficult to distinguish between two different samples in the data set.

Finally, existing iterative methods such as \cite{10.1121/10.0009851} and ML methods like \cite{waqasahmed1, waqasahmed2} optimize their model parameters using solely $Loss_{FarField}$, which makes the forward pass essential for the methods.
Our observations confirm that, in contrary, the shape reconstruction process in IASPs does not have to depend on the forward pass to produce high-quality results. The inverse network, a convolutional neural network equipped with non-linear activation functions, it is able to successfully learn the severely ill-posed and non-linear mapping between the scattering information and the scatterer shapes.

\subsection{\label{subsec42} Case 2: ShapeNet}
\begin{figure}[!htbp]
    \centering
    \includegraphics[width=0.5\textwidth]{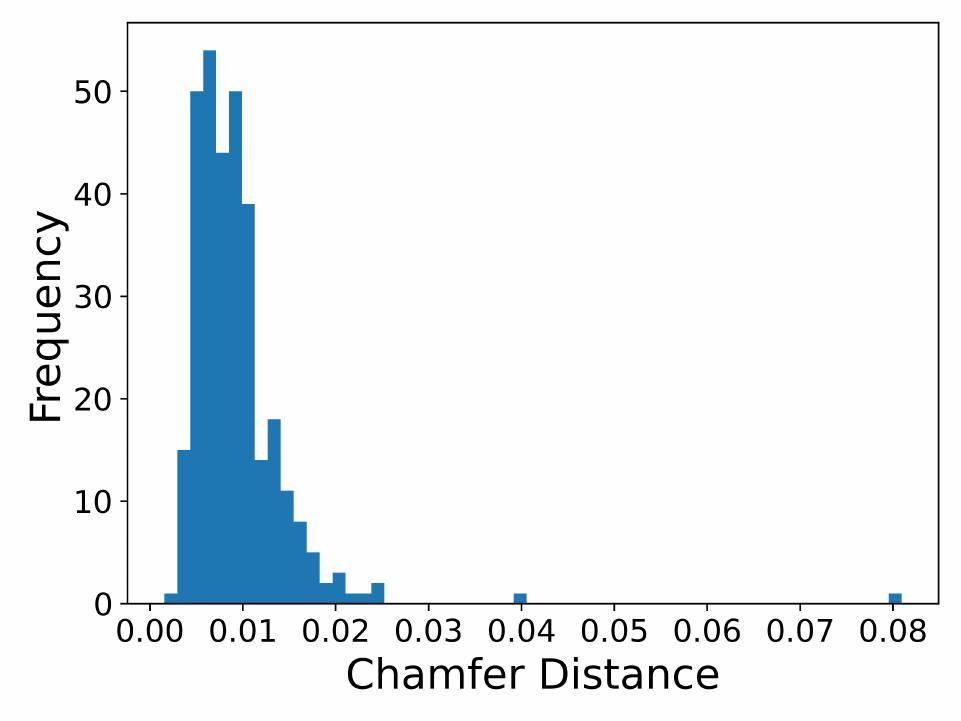}
    \caption{Error distribution of shape reconstruction results for 320 test samples from the \textit{cars} class.}\label{fig:Figure11}
\end{figure}

\begin{figure}
    \includegraphics[width=0.5\textwidth]{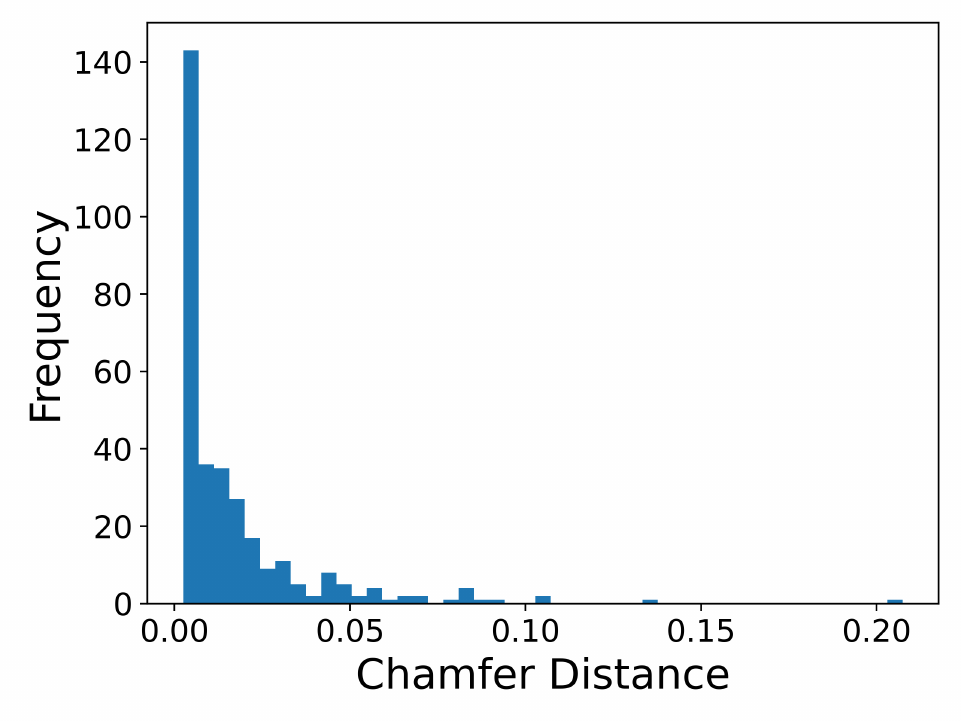}
    \caption{Error distribution of shape reconstruction results for 320 test samples from the \textit{airplanes} class.}\label{fig:Figure12}
\end{figure}

\begin{figure*}[hbtp!]
    \centering
    \includegraphics[width=0.8\textwidth]{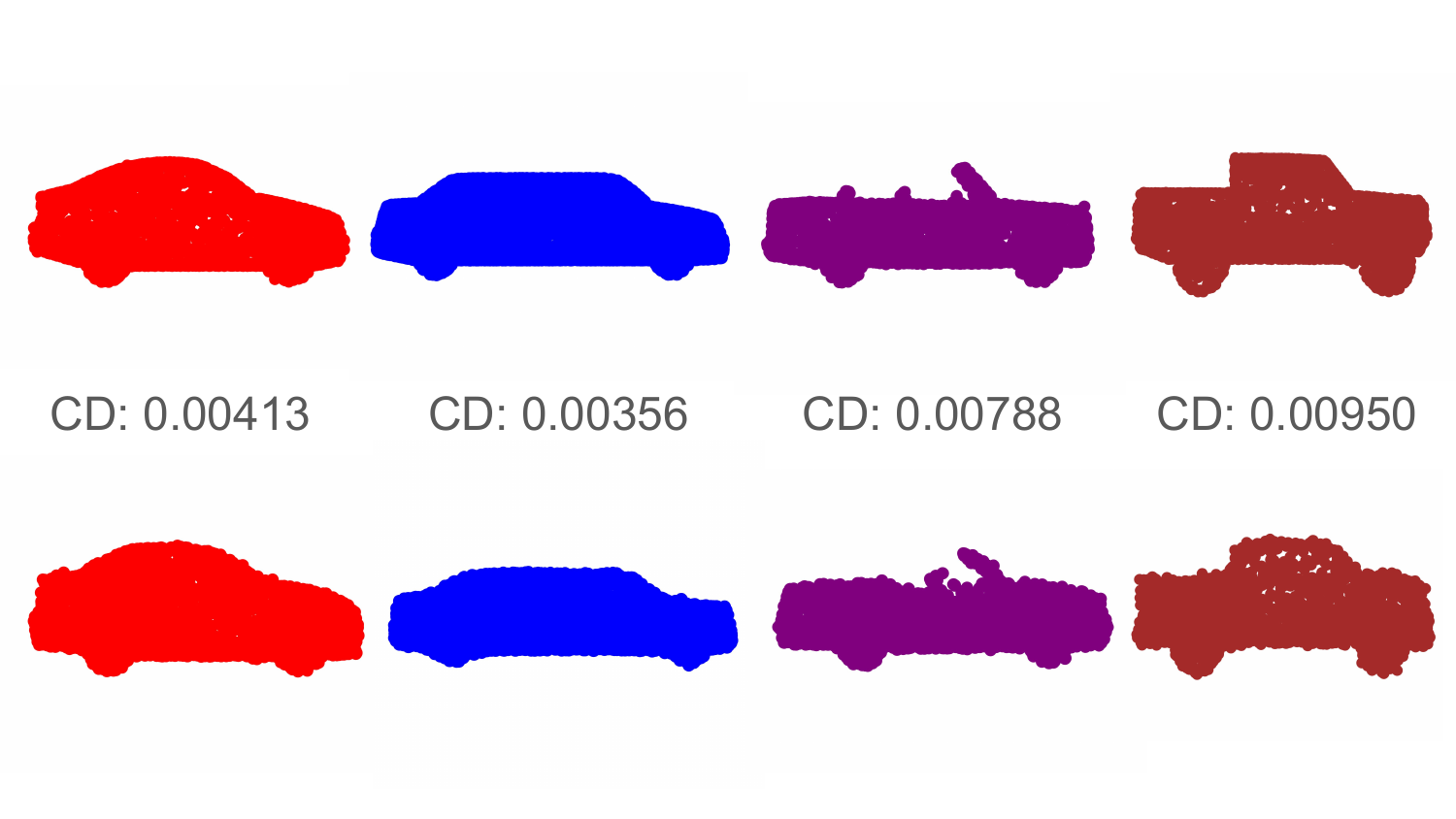}
    \caption{Ground-truth (top) and reconstructed (bottom) scatterer point-clouds of test samples from the \textit{cars} class. The reconstruction error in Chamfer distance (CD) is given for each sample.}
    \label{fig:Figure13}
\end{figure*}
\begin{figure*}[hbt!]
    \centering
    \includegraphics[width=0.8\textwidth]{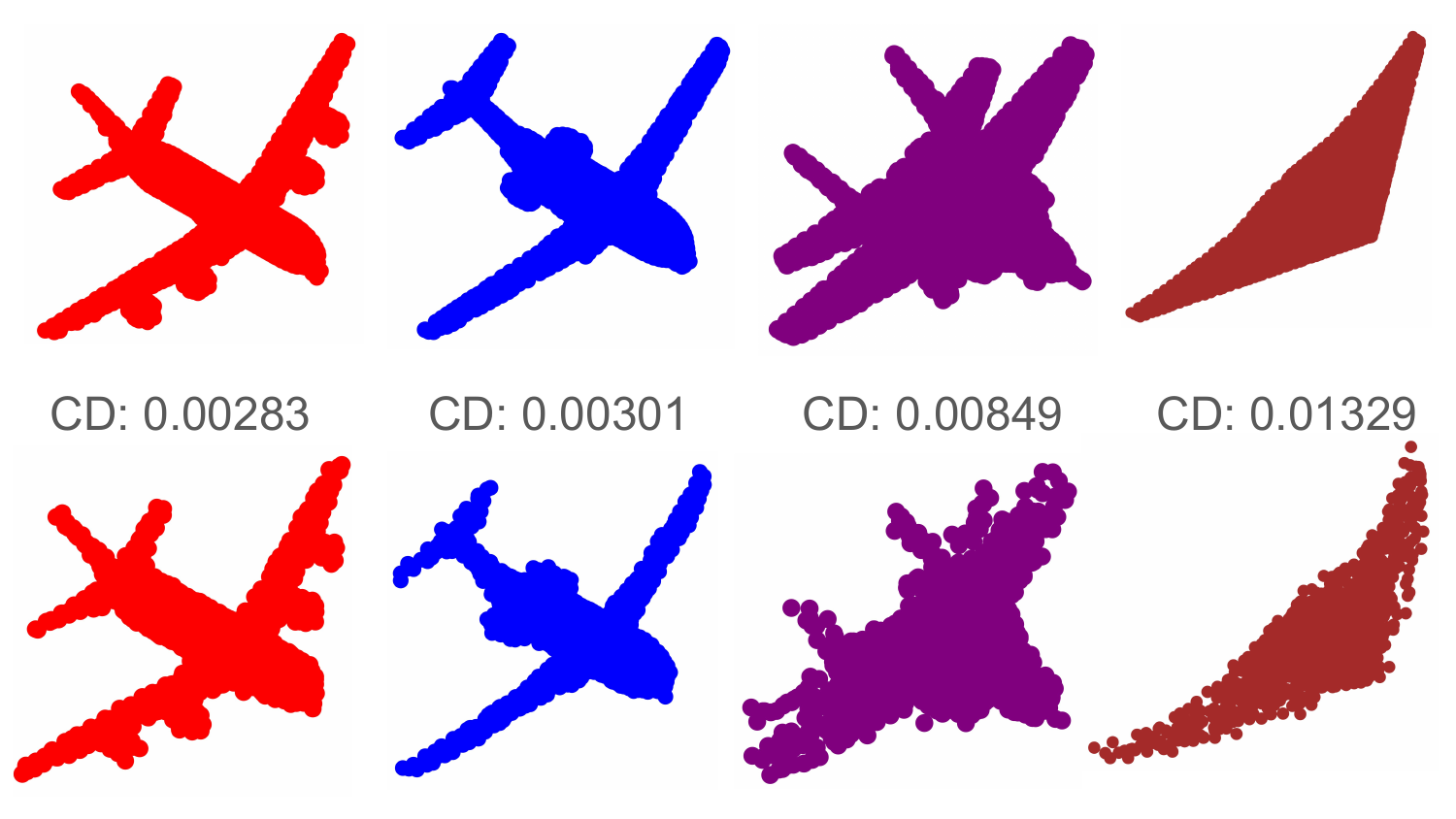}
    \caption{Ground-truth (top) and reconstructed (bottom) scatterer point-clouds of test samples from the \textit{airplanes} class. The reconstruction error in Chamfer distance (CD) is given for each sample.}
    \label{fig:Figure14}
\end{figure*}

While the random particle data set provides a practical way of testing the proposed method, it doesn't contain any common objects from benchmark computer vision data sets that would allow us to make a more meaningful evaluation. The \textit{airplane} and \textit{cars} classes of the ShapeNet data set, help us to address this issue. Both classes contain very different shapes that have distinct complex features. Also, under the light of the results obtained in the previous section, we optimize the framework using $Loss_{shape}$, bypassing the forward step completely. Figure \ref{fig:Figure13} and Figure \ref{fig:Figure14} show the reconstruction results for test samples drawn from the \textit{cars} and \textit{airplane} classes respectively. Note that the camera is rotated to a specific angle for each case, to demonstrate the differences between the reconstructions and the ground-truth data more effectively. The left column contains the ground-truth point clouds of the scatterers and the right column contains the reconstructed point clouds, by the proposed framework. We intentionally pick samples that belong to different subclasses, having either significant local and/or global structural differences. The framework is able to learn most global and local features, as it can be seen from the figures. 
For the \textit{cars} class, we can easily see that a limousine (blue), a convertible (purple) and a truck (brown), which all have distinct features, are successfully reconstructed. However, we also observe subtle errors in the reconstructions. For example the number of seats in the convertible are not predicted correctly. Also, the corners of the roof in the truck example are not as sharp. These kind of reconstruction errors are observed throughout the test data set. However, as the framework is data-driven, these imperfections are expected and strongly depend on the training data too. Figure \ref{fig:Figure11} shows the error (Chamfer Distance) distribution of 320 test samples from each data set. As it can be seen from the figure, most reconstruction errors are accumulated around $0.01$. 

With the \textit{airplanes} class, we observe much more complex features and diversity amongst the scatterers. As it can be seen from Figure \ref{fig:Figure14}, the framework is able to successfully differentiate between the number and location of the jet propellers, and the global structure of the different aircraft. Again, we observe a loss of density and accuracy in the fighter jet (purple) and stealth bomber (brown) reconstructions. This is partly due to the fact that half of the data set consists of commercial airliners, which is also reflected in the error distribution in Figure \ref{fig:Figure12}, where the lower errors mostly belong to commercial airliner reconstructions, and there are much less examples of other aircraft types. Still, the framework is able to capture the overall global and local properties of the shape, like the tail-wings and sharp wing features in the fighter jet reconstruction. Figure \ref{fig:Figure12} shows the error distribution of the test samples. Again, the errors are mostly accumulated around $0.01$.

 Lastly, we evaluate the performance of the proposed method, relative to the iteration time of the numerical solver utilized in \cite{alsnayyan2022laplace}. To this end, we report the execution time of the numerical forward scattering solver, for the airplane object in the top row of Figure \ref{fig:Figure1} at $750$ Hz. The forward solver takes $954$ seconds to complete on a single Intel(R) Xeon(R) Gold 6148 CPU. This would mean that a single iteration of the inverse shape optimization procedure for this airplane would approximately take $954$ seconds. The proposed method, on the other hand, is able to compute the predictions for $322$ airplane objects in the test data, in $18$ seconds ($0.056$ sec/airplane). This is several orders of magnitude faster, rendering the proposed framework appealing to a wide range of practical applications.

\section{\label{sec:5} Conclusion} 

In this paper, we have demonstrated ISSRNet, a deep learning framework for solving the 3D inverse acoustic scattering for shape reconstruction problem. Using a convolutional neural network, ISSRNet encodes the scattering far-field data into a latent space. Then it maps this scattering latent space, to the 3D shape latent space, which is learned by a 3D variational auto-encoder. ISSRNet only requires data from a single incident wave, at a single frequency and performs orders of magnitudes faster than a traditional iterative method, while still capturing both global and local shape details about complex scatterers. Moreover, in contrast to existing iterative and machine learning solutions, ISSRNet does not depend on the forward solution for the scattering problem. We evaluate the proposed framework on both a synthetic random 3D shape data set with a high amount of random surface variation; as well as the \textit{cars} and \textit{airplanes} classes of the popular 3D shape data set ShapeNet. As is evident, the results are extremely promising, while leaving room for improvement to capture finer grained details. These improvements include, using multiple frequencies to interrogate the object, using multiple incident field data, better shape descriptors through a more physics-aware encoder and/or loss function and so on. These will be topics that will be discussed in subsequent papers. 

\section{\label{sec:6} Acknowledgements}
This work was supported by the National Science Foundation (NSF) under awards OAC-1845208. Computational resources were provided by the High Performance Computing Center (HPCC) at Michigan State University.

\bibliographystyle{IEEEtran}
\bibliography{references}

\end{document}